\documentclass[runningheads]{llncs}
\usepackage{graphicx}

\usepackage{tikz}
\usepackage{comment}
\usepackage{amsmath,amssymb} 
\usepackage{color}
\usepackage{graphicx}
\usepackage{amsmath}
\usepackage{amssymb}
\usepackage{booktabs}
\usepackage{bbold}
\usepackage{array}
\usepackage{mathtools}
\usepackage{multirow}

\usepackage{algorithm}
\usepackage{algpseudocode}

\usepackage[accsupp]{axessibility}  

\DeclareMathOperator*{\argmin}{arg\,min}
\newcommand{\ours}{EvAC3D}
\newcommand{\dataset}{MOEC-3D}
\newcolumntype{?}{!{\vrule width 1pt}}

\DeclareMathOperator{\SurfacePoint}{{}^{w}\textbf{X}(t)}
\DeclareMathOperator{\SurfaceNormal}{{}^{w}\textbf{n}(t)}

\DeclareMathOperator{\CamCenter}{{}^{w}\textbf{p}_c(t)}
\DeclareMathOperator{\CamVector}{{}^{c}\textbf{x}(t)}

\DeclareMathOperator{\CamRotation}{{}^{w}\textbf{R}_c(t)}

\usepackage[pagebackref,breaklinks,colorlinks]{hyperref}
\usepackage{orcidlink}
\usepackage[capitalize]{cleveref}
\crefname{section}{Sec.}{Secs.}
\Crefname{section}{Section}{Sections}
\Crefname{table}{Table}{Tables}
\crefname{table}{Tab.}{Tabs.}

\begin{document}
\pagestyle{headings}
\mainmatter
\def\ECCVSubNumber{5100}  

\title{EvAC3D: From Event-based Apparent Contours to 3D Models via Continuous Visual Hulls} 


\titlerunning{EvAC3D}

%
\author{Ziyun Wang*\orcidlink{0000-0002-9803-7949} \and
Kenneth Chaney*\orcidlink{0000-0003-1768-6136} \and
Kostas Daniilidis\orcidlink{0000-0003-0498-0758}}
\authorrunning{Wang et al.}
%
\institute{University of Pennsylvania, Philadelphia PA 19104, USA}
\newcommand{\claude}[1]{{\color{red} Claude: #1}}
\maketitle

\begin{abstract} 

3D reconstruction from multiple views is a successful computer vision field with multiple deployments in applications. State of the art is based on traditional RGB frames that enable optimization of photo-consistency cross views. In this paper, we study the problem of 3D reconstruction from event-cameras, motivated by the advantages of event-based cameras in terms of low power and latency as well as by the biological evidence that eyes in nature capture the same data and still perceive well 3D shape. 
The foundation of our hypothesis that 3D-reconstruction is feasible using events lies in the information contained in the occluding contours and in the continuous scene acquisition with events.
We propose Apparent Contour Events (ACE), a novel event-based representation that defines the geometry of the apparent contour of an object. We represent ACE by a spatially and temporally continuous implicit function defined in the event x-y-t space. Furthermore, we design a novel continuous Voxel Carving algorithm enabled by the high temporal resolution of the Apparent Contour Events. To evaluate the performance of the method, we collect \dataset, a 3D event dataset of a set of common real-world objects. We demonstrate \ours{}'s ability to reconstruct high-fidelity mesh surfaces from real event sequences while allowing the refinement of the 3D reconstruction for each individual event. The code, data and supplementary material for this work can be accessed through the project page: \url{https://www.cis.upenn.edu/~ziyunw/evac3d/}.

\end{abstract}
\section{Introduction}
\label{scn:intro}
\begin{figure}
    \centering
    \includegraphics[width=\textwidth]{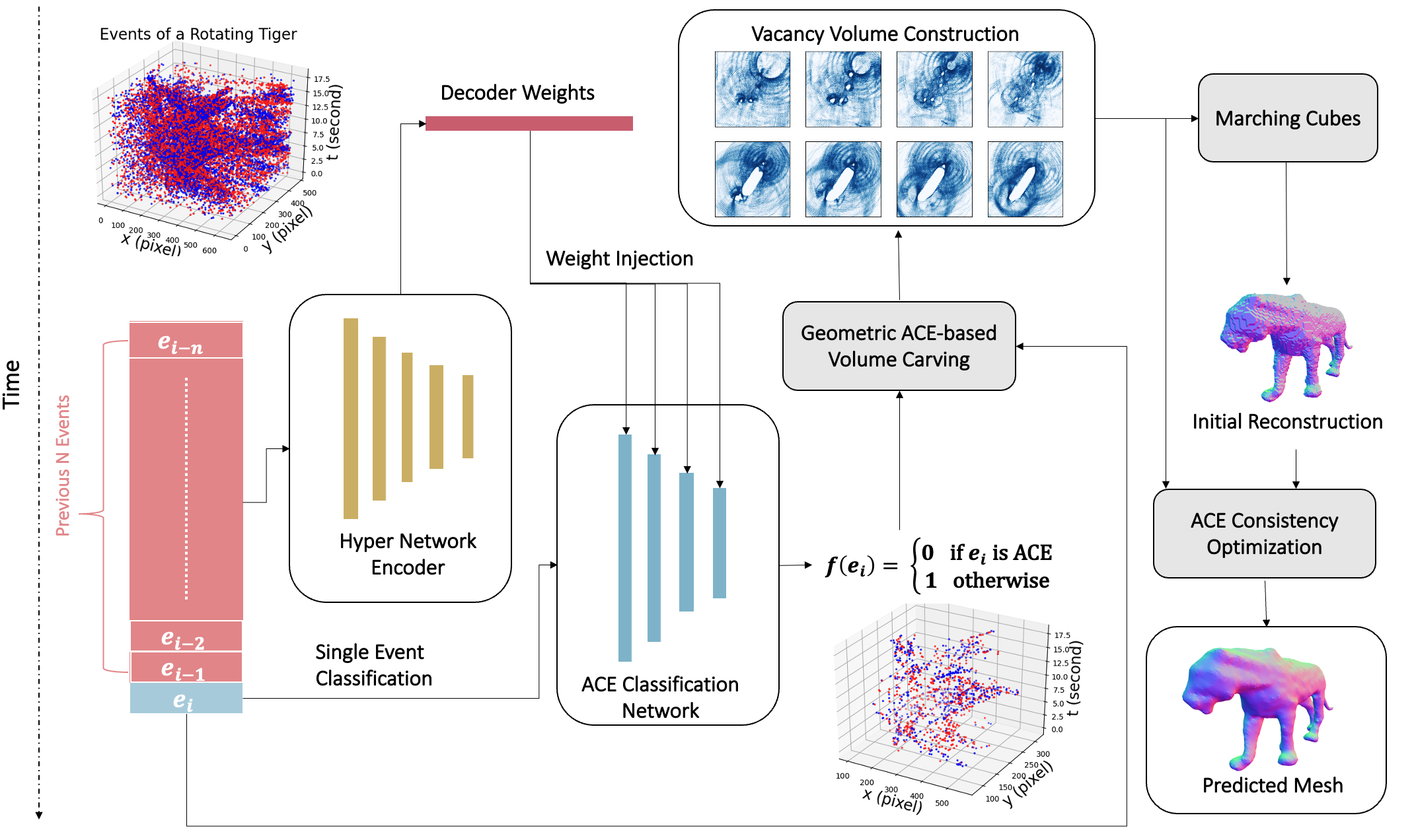}
    \caption{\textbf{\ours{}} Pipeline. We use the previous N events as conditional information to predict the label for the current event. A hyper network is used to inject the conditional information into the decoding classifier. The predicted label is then passed into a geometry-based volume event carving algorithm.}
    \label{fig:figure1}
\end{figure}

Traditional 3D reconstruction algorithms are frame-based because common camera sensors output images at a fixed frame rate. The fixed frame rate assumption challenges researchers to develop complex techniques to handle undesirable situations due to discontinuity between frames, such as occlusions. Therefore, recovering the association between views of the same object has been an essential problem in 3D reconstruction with a single camera. Such challenges fundamentally arise from the discrete time sampling of visual signals, which forces vision algorithms to recover the missing information between views. However, these problems do not exist naturally in biological systems because visual signals are encoded as a stream of temporally continuous spikes. Continuous encoding tremendously benefits humans and animals in many tasks, including estimating the 3D geometry of an object. The question is: \textit{can a computer vision algorithm do better if it sees the same continuous world as humans do?}

\begin{figure}
    \centering
    \includegraphics[width=0.7\textwidth]{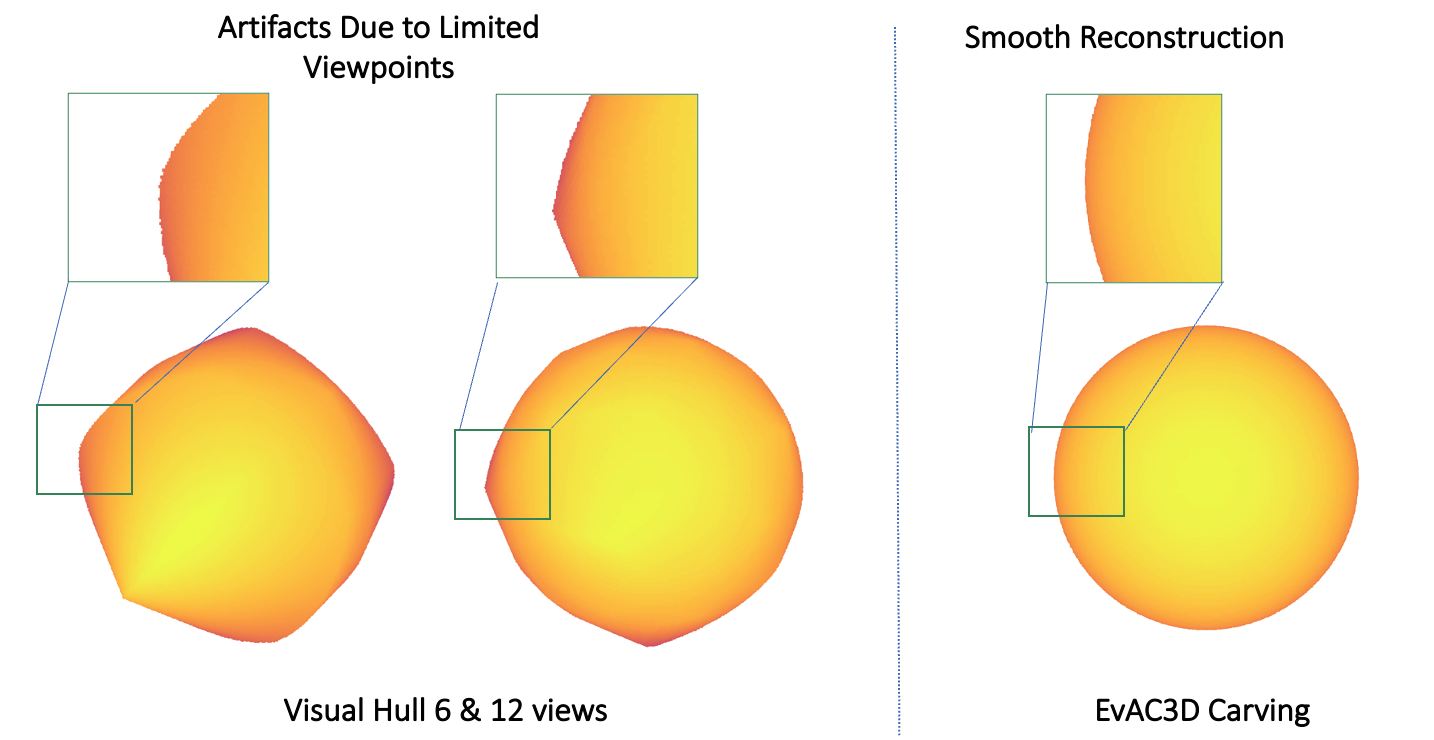}
    \caption{Reconstruction of a sphere with visual hull (6 and 12 frames) and with \ours{} reconstruction on simulated events. The 12-view visual hull method uses roughly the same number of operations as \ours{}.}
    \label{fig:sphere}
\end{figure}

In this work, we seek the answer to this question by developing a novel algorithm for bio-inspired event-based cameras. Event-based cameras are novel visual sensors that generate a continuous stream of events triggered by the change of the logarithm of the light intensity. The events are asynchronously generated without temporal binning; therefore, the high-resolution temporal information can be completely recovered from each event with minimal discontinuity. Additionally, the individual pixels of the camera do not have a global shutter speed, which gives the camera extremely high dynamic range. Due to the high dynamic range and high temporal resolution of event cameras, they have become an ideal choice for understanding fast motions. For 3D reconstruction, traditional cameras operate on a fixed frame rate. 
For image-based visual hull methods, the limited number of views means the smooth surfaces of the object cannot be properly reconstructed, which can be seen from the sphere reconstruction example in Figure~\ref{fig:sphere}. 

Ideally, one can expect to directly perform incremental updates to the geometry of an object at the same high temporal resolution as events. To this end, we propose a 3D reconstruction pipeline that directly predicts a mesh from a continuous stream of events assuming known camera trajectory from a calibrated camera. We introduce a novel concept of \textbf{Apparent Contour Events} to define the boundary of an object in the continuous x-y-t space. Through Apparent Contour Events, we incrementally construct the  function of a 3D object surface at the same high temporal resolution as events. Here is a list of our main contributions:
\begin{itemize}
    \item We introduce a novel event concept of \textbf{Apparent Contour Events} that relates high-speed events to the tangent rays of the 3D surface to the viewpoint. 
    \item We propose a learning pipeline to predict which events are Apparent Contour Events without manual annotation but using 3D models of known objects. We propose a novel event-based neural network with point-based decoding to classify the Apparent Contour Events.
    \item We present a continuous algorithm to reconstruct an object directly from a stream of events. The algorithm can accurately reconstruct objects with complex geometry in both synthetic and real environments.
    \item We collect \dataset, a high-quality real 3D event dataset for evaluating the performance of 3D reconstruction techniques from events that provides events, ground-truth 3D models, and ground-truth camera trajectories.
\end{itemize}

\section{Related Work}
\textbf{3D Reconstruction with Event Cameras} 
Due to the asynchronous and sparse nature of the event sensors, 3D reconstruction algorithms cannot be directly applied. Most current work in event-based 3D reconstruction uses a stereo pair of cameras~\cite{kim2016real,carneiro2013event,zhou2018semi,zhu2018realtime}. The time coincidence of the events observed from a synchronized pair of cameras is used for stereo matching. These methods work in situations where multiple calibrated cameras are used synchronously. Zhu et al.~\cite{zhu2018realtime} construct a cost volume based on warping using multiple disparity hypotheses. Carneiro et al.~\cite{carneiro2013event} use time incidence between two synchronized event streams to perform stereo matching. Chaney et al.~\cite{chaney2019learning} use a event single camera in motion to learn the 3D structure of the world. E3D~\cite{baudron2020e3d} attempts to directly predict meshes from multi-view event images. This method is trained and mainly evaluated on synthetic data due to the large amount of 3D  data needed for training. EMVS~\cite{rebecq2018emvs} adopts an event-based ray counting technique. Similar to our method, EMVS treats individual events as rays to take advantage of the sparse nature of the event data. In Section~\ref{sec:carving}, we show how sparse processing can be extended further to work with only a particular type of events that contain rather rich geometric information.\\\\
\textbf{3D Reconstruction from Occluding Contours.} Reconstruction from the perspective projection of a geometric surface has been extensively studied in classical computer vision. Among different geometric representations used in such problems, apparent contour representation is most relevant to our work. Apparent contours, or extreme boundaries of the object, contain rich information about a smooth object surface. Barrow et al.~\cite{barrow1981interpreting} argue that surface orientations can be directly computed from a set of apparent contours in line drawings. Cipolla et al.~\cite{cipolla1992surface} propose the theoretical framework from reconstructing a surface from the deformation of apparent contours. Based on the idea of contour generator~\cite{marr1977analysis}, the projection of the apparent contours onto the image plane are used as tangent planes to the object. Furthermore, structure and motion can be simultaneously recovered from apparent contours. Wong et al.~\cite{wong2001structure} propose to solve the camera poses and 3D coordinates of ``frontier points", the intersection of the apparent contours in two camera views. A circular motion with a minimum of 3 image frames is assumed to solve the optimization problem.\\\\
\textbf{Visual Hull.} Visual hull is used to reconstruct 3D objects through Shape-From-Silhouette (SFS) techniques \cite{baumgart1974geometric,laurentini1991visual}. Information from multiple views are aggregated into a single volume through intersection of the projective cones described by the silhouette at each view. Voxel grid and octrees \cite{hornung13auro,szeliski1993rapid} are commonly used as discretized volumetric representations. SFS methods are particularly susceptible to false-negative labels (labeling an interior point as an exterior point).


\begin{figure}
\centering
    \includegraphics[width=\textwidth]{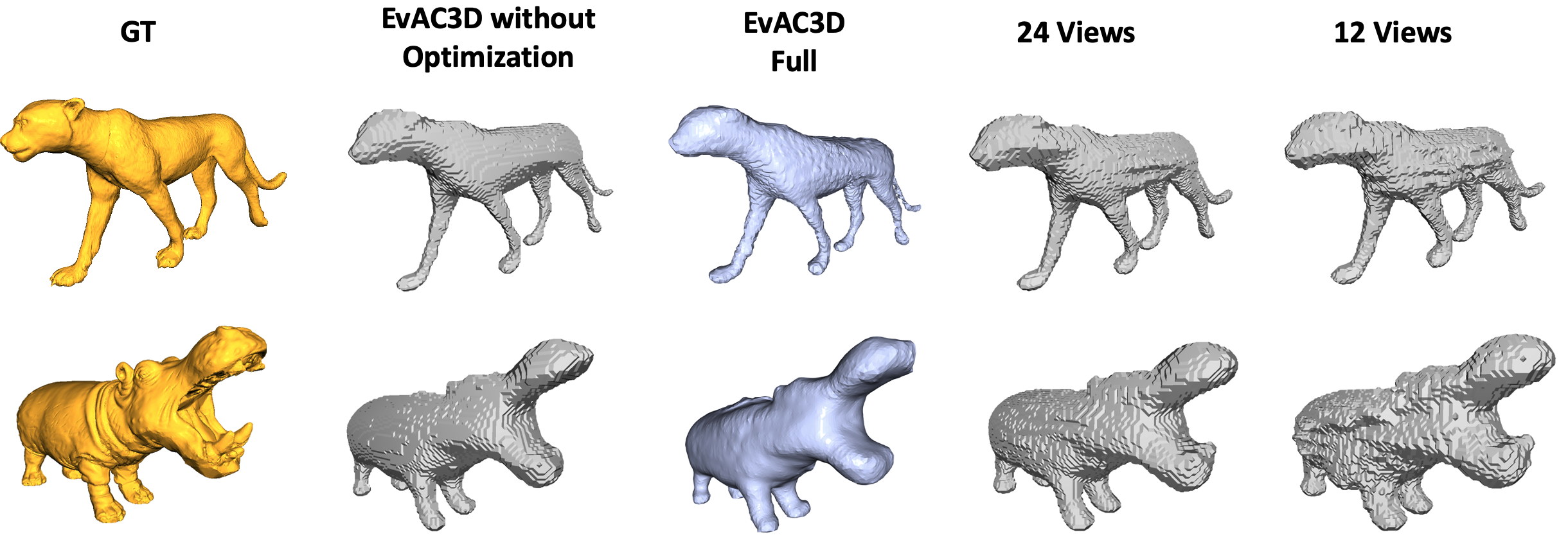}
  \caption{Qualitative comparisons between \ours and mask based carving of 12 and 24 views respectively. Cheetah, hippo, and elephant were chosen as a subset of the animal scans.}\label{fig:evac_carving}
\end{figure}

\section{Method}
In this section, we explain how a continuous stream of events can be used to reconstruct the object surface. We divide the pipeline into two stages: \textbf{Apparent Contour Event Extraction} and \textbf{Continuous Volume Carving}.

\subsection{Apparent Contour Event (ACE)}
The main challenge in reconstructing objects from events is finding the appropriate geometric quantities that can be used for surface reconstruction. In frame-based reconstruction algorithms, silhouettes are used to encode the rays from the camera center to the object. However, computing silhouettes requires integrating frame-based representations, which limits the temporal resolution of the reconstruction updates. Additionally, since events represent the change in log of light intensity, events are only observed where the image gradients are nonzero. Therefore, one would not observe enough events on a smooth object surface. These two facts combined make traditional silhouettes non-ideal for events. To address these two shortcomings, we introduce \textbf{Apparent Contour Events (ACE)}, a novel representation that encodes the object geometry while preserving the high temporal resolution of the events.

\begin{figure}


\includegraphics[width=\textwidth]{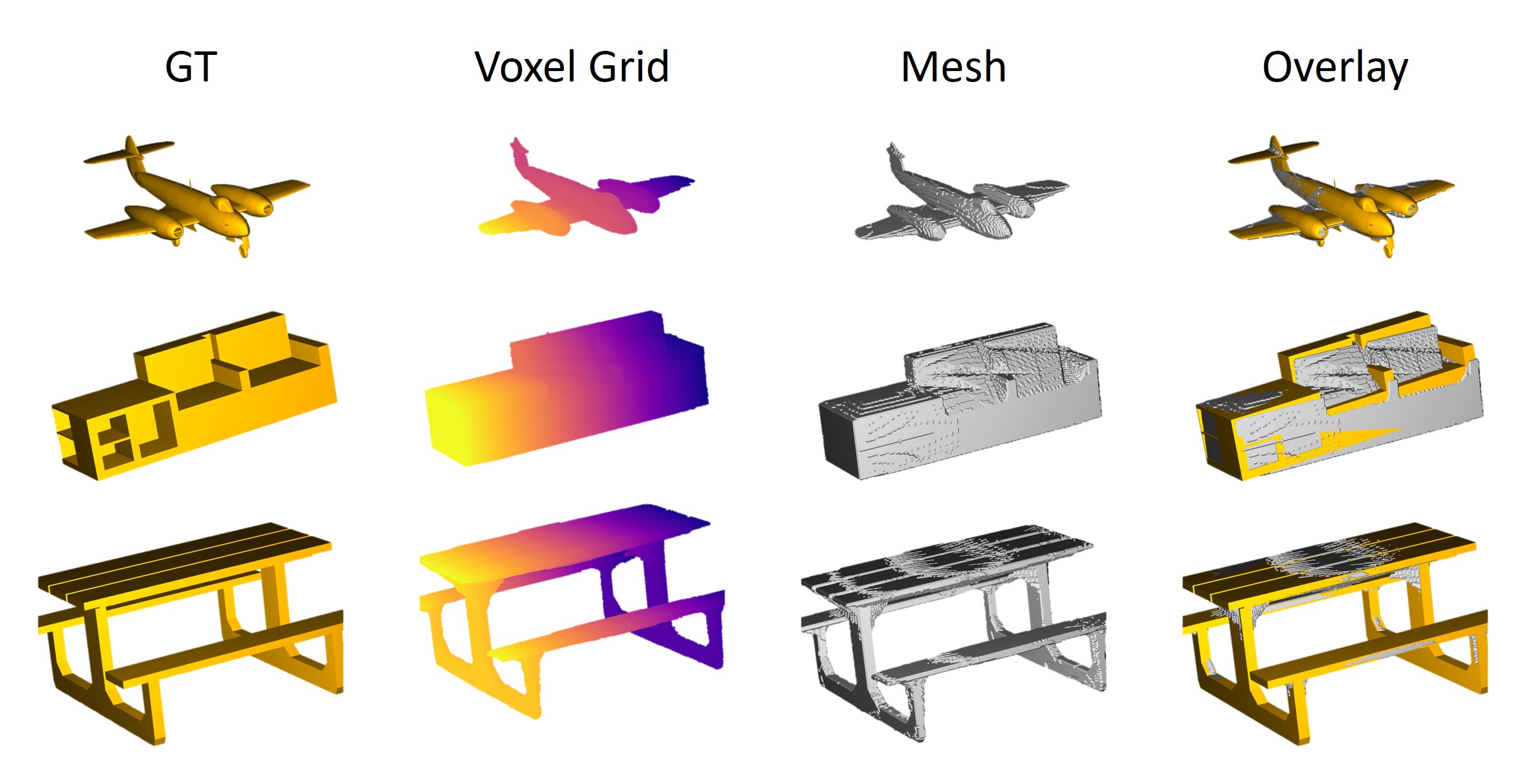}

\caption{Qualitative evaluations from ShapeNet using \ours{} on three categories of objects.}\label{fig:shapnet_results}
\end{figure}

Geometrically, the generator of occluding contours on image planes is constrained by a ray-surface intersection and tangency \cite{giblin1987reconstruction}. A smooth surface $\mathcal{S}$ with well defined surface normals at each point has an occluding contour generator for each camera center $\mathbf{{}^{w}p_{c}}$. The contour generator is composed of image rays that intersect the $S$ at exactly one point $\mathbf{{}^{w}X}$. A surface point $\mathbf{{}^{o}x}$ with normal $\mathbf{{}^{w}n}$ is included in the contour generator for the camera center $\mathbf{{}^{w}p_c}$ if for an image ray $\mathbf{{}^{w}v}$ the ray-surface intersection and tangency constraints hold \cite{giblin1987reconstruction}:
\begin{align}
    \lambda \mathbf{{}^{o}v} + \mathbf{{}^{w}p_c} = \mathbf{{}^{w}X} \\
    \mathbf{{}^{w}\!n^\intercal} (\mathbf{{}^{w}\!X} - \mathbf{{}^{w}p_c}) = 0
\end{align}
We define Apparent Contour Events formally. ACEs are events that meet the ray-surface intersection and tangency constraints \cite{giblin1987reconstruction}. Since each event can contain a potentially unique timestamp, the constraints must be thought of in continuous time, as opposed to indexible on a per frame basis. An event $e_i$ generates an image ray $\CamVector$ at a camera center $\CamCenter$. $e_i$ is an ACE if for some point $\SurfacePoint$ on the surface $\mathcal{S}$:
\begin{align}
    \SurfaceNormal^\intercal ( \SurfacePoint - \CamCenter) &= 0\\
    \lambda(t) \CamRotation \CamVector + \CamCenter &= \SurfacePoint
\end{align}
Intuitively, ACE can be seen as the set of events $e_i = \{x_i, y_i, t_i, p_i\}$ that belong to the active contour of the object at time $t_i$. Due to the contrast between the active contour of an object with the background, a significant number of events are generated around the contour. Unlike silhouettes, which require filling in holes on the ``eventless" areas of an integrated image, an ACE is defined purely on events. Traditional algorithms are limited by the frame rate of the input images. Projecting rays from only through the contours produces far fewer intersections of the rays. With events, we can shoot a ray for each event, which continuously refine the geometry around the active contour, as shown in Figure~\ref{fig:figure1}. To fully take advantage of the continuous nature of the events, we design a novel continuous volume carving algorithm based on single events, as described in Section~\ref{scn:carve}.

\begin{figure}
    \centering
     \includegraphics[width=0.9\textwidth]{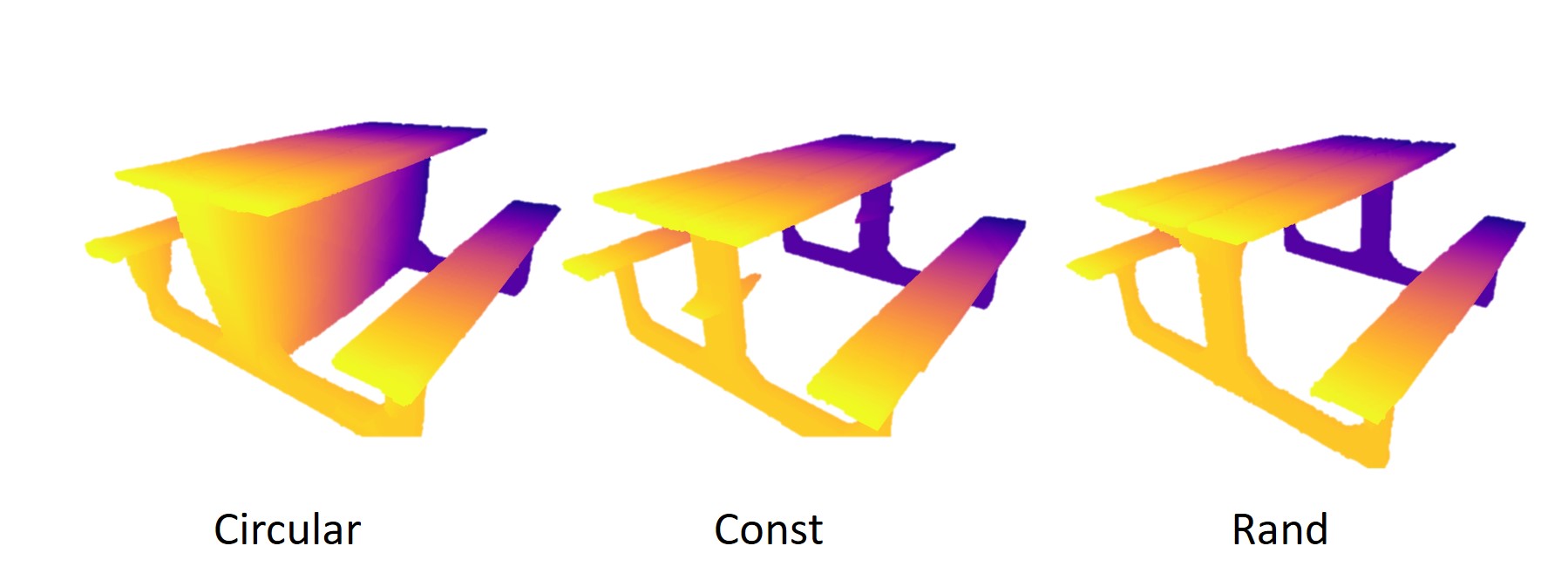}
    \caption{Comparison of different trajectories in simulation with ShapeNet. The circular and octahedral trajectories only move around major axes missing some contours that would improve the carving results. In comparison, the random trajectory samples evenly across the sphere providing more unique viewpoints.}
    \label{fig:viewpoint}
\end{figure}

\subsection{Learning Apparent Contour Events}
We formulate identification of Apparent Contour Events (ACEs) as a classification problem. In other words, the network learns a function $F_{E_{t_i}}$, which maps an event to whether it is an ACE conditioned on the history of events $E_{t_i}$. For an event $e_i = \{x_i, y_i, t_i, p_i\}$, we encode the past $N$ events using a function $\theta$ as a $K$ dimensional latent vector $C_i \in \mathbf{R}^K$, where $K$ is a hyperparameter.
\begin{align}
    C_{i} &= g_{\phi}(\{e_j \coloneqq (x_j, y_j, t_j, p_j)\}): j > \max(i - N, 0)
\end{align}
$N$ is a hyperparameter that specifies the history of events as the conditional input to the classification problem. The ACE classification problem is modeled as a function that maps from the latent code and an event to the probability that it is an ACE:
\begin{align}
    q_i &= f_{\theta}(e_i, C_{i}) \\
    e_i &\coloneqq (x_i, y_i, t_i, p_i) \\
    &q_i \in [0, 1]
\end{align}
We use a neural network to parameterize function $g_\phi$ and $f_\theta$. Note that $g_\phi$ takes a list of $N$ events. In practice, we use an event volume~\cite{zhu2019unsupervised} to encode past events.
\begin{align}
E(x,y,t)=&\sum_{i} p_i k_b(x-x_i)k_b(y-y_i)k_b(t-t^*_i)\label{eq:event_volume}
\end{align}

We chose this representation because the values in such volumes represent the ``firing rate" of visual neurons in biological systems, which preserves valuable temporal information. The temporal information is needed because labeling ACEs requires the network to predict both where the contours are in the past and how they move over time. To supervise the ACE network, we jointly optimize the encoder and the event decoder using a Binary Cross-Entropy loss directly on the predicted event labels.
\begin{align}
\mathcal{L}_{c} =\frac{1}{N} \sum_{i=1}^{N} \mathcal{L}_{bce}(f_{\theta}(e_i, C_{i}), \hat{q_{i}}))
\end{align}
Here $\hat{q}_i$ is the ground truth event label for $e_i$. In practice, the labels are extremely imbalanced especially in either low light conditions (high noise to signal ratio) or scenes where other objects are moving as well. For training, we equally sample half positive and half negative events to help overcome the imbalance of labels.
\begin{algorithm}
    \caption{Event Carving Algorithm}\label{algo:event_carving}
     \hspace*{\algorithmicindent} \textbf{Input} $V$ volume initialized to zero \\
     \hspace*{\algorithmicindent} \textbf{Input} $E$ active contour events \\
     \hspace*{\algorithmicindent} \textbf{Input} $\CamRotation, \CamCenter$ camera trajectories
    \begin{algorithmic}[1]
        \Procedure{CarveEvents}{$V$, $E$, $\CamRotation$, $\CamCenter$}
        \For{$i\gets 1, |E|$}
            \State $(x_i, y_i, t_i, p_i) \gets E_i$
            \State ${}^{V}T_{C(t_i)} \gets {}^{V}T_{W}{}^{W}T_{C(t_i)}$
            \State $O_i \gets {}^{V}\textbf{R}_{W} {}^{w}\textbf{p}_c(t_i) + {}^{V}\mathbf{t}_W$
            \State $D_i \gets {}^{V}\textbf{R}_{W} {}^{w}\textbf{p}_c(t_i) {}^{w}\textbf{x}_c(t_i)$
            \State $V_i \gets bresenham3D( O_i, D_i, bounds(V) )$
            \State $V[V_i] += 1 $
        \EndFor
        \State \textbf{return} $V$
        \EndProcedure
    \end{algorithmic}
\end{algorithm}
\paragraph{Architecture} To enable classification of individual events, we adopt an encoder-decoder architecture where the decoder maps event coordinates to probabilities. These types of architectures are widely used in learning-based single-view 3D reconstruction methods. Mapping approaches such as AtlasNet~\cite{groueix2018papier} and implicit approaches (Occupancy Networks~\cite{mescheder2019occupancy}, DeepSDF~\cite{park2019deepsdf}) all use variants of this architecture. In our experiments, we find the decoder part of the network has more weight in the overall mapping performance. Rather than taking fixed-sized latent vector code, we inject the conditional information directly into the weights of the decoder, following~\cite{ha2016hypernetworks,mescheder2019occupancy,wang2020geodesic,wang2020surface,mitchell2019higher}. We use the Conditional Batch Normalization to inject the encoding of the prior events into the Batch Normalization layers of the decoder network. The architecture of the network is illustrated in Figure~\ref{fig:figure1}. The training details and hyperparameters of the network can be found in the Supplementary Material.

\subsection{Event Based Visual Hull}
\label{sec:carving}
In frame-based shape from silhouette and space carving approaches, the goal is to recover the visual hull, defined as the intersection of visual cones that are formed by the apparent contour in each frame. A better definition, though following the original definition by Laurentini \cite{laurentini1991visual} would be the largest possible volume consistent with the tangent rays arising from apparent contour events. 
 The visual hull is always a superset of the object and a subset of the convex hull of the object. Due to the continuity of the camera trajectory and the high temporal sampling of events, we expect the obtained visual hull to be tighter to the object than the visual hull obtained from a sparse set of viewpoints that might be closer to the convex hull of the object. 
 
\subsubsection{Continuous Volume Carving}
\begin{figure}
    \centering
    \label{fig:sphere_and_auc}
    \includegraphics[width=0.9\textwidth]{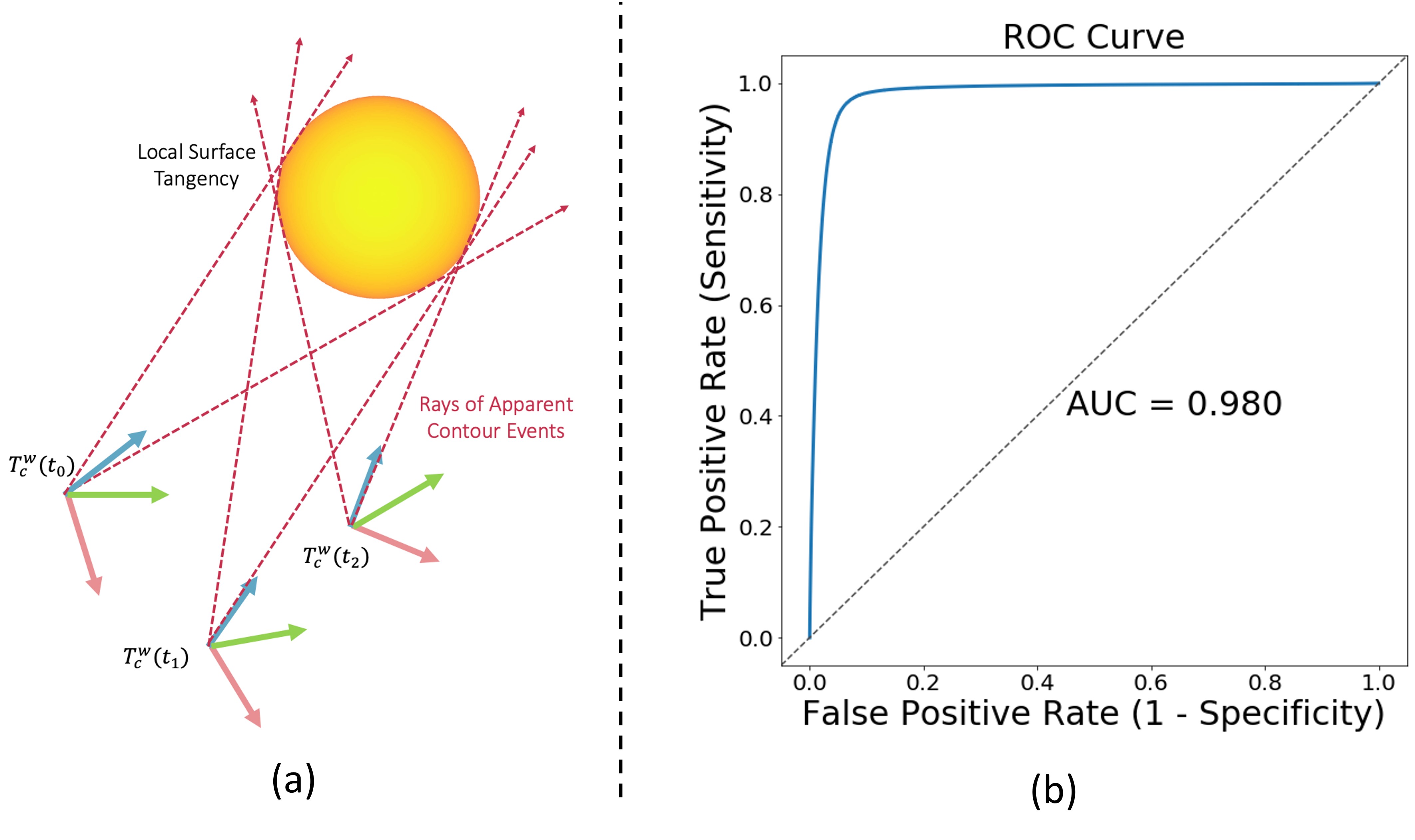}
\caption{(a): illustration of carving based on Apparent Contour Events. (b): ROC curve of ACE classification network.}
\end{figure}
\label{scn:carve} 
provides smooth continuous incremental changes to the carving volume. This is accomplished by only carving updates through the use of ACEs. This creates a more computationally efficient as shown in Table~\ref{tab:compute}.

ACEs are defined by the tangent rays to the surface at any given positional location. The time resolution of event based cameras provide ACEs that are from continuous viewpoints through the trajectory of the camera. These continuous viewpoints, $C(t)$, are from around the object in the world frame, $W$. Projecting an individual ACE, into the voxel grid coordinate system gives us a ray with origin, ${}^{V}R_{W} \CamCenter + {}^{V}\mathbf{t}_W$, and direction, ${}^{V}R_{W} \CamRotation \CamVector$. The ray in the voxel grid coordinate system allows us to project through the volume, which is illustrated in Figure~\ref{fig:sphere_and_auc} (a). To efficiently traverse this volume, a 3D Bresenham algorithm is used to produce a set of voxel coordinates, $\mathcal{V}_{i} \in \mathbb{Z}^3$, along the ray. All voxels in $\mathcal{V}_{i}$ are incremented. The interior of the object is left empty, as the rays trace along the continuous surface of the object. This can be seen at the bottom of Figure~\ref{fig:figure1}. Algorithm~\ref{algo:event_carving} covers the process of generating updates to the voxel grid for every event individually. The object's mesh is then extracted from the volume (algorithm in the supplementary material) and optimized.



\subsection{Global Mesh Optimization}
The mesh reconstructed from volume carving can be affected by noise either from pose estimation or sensor noise. Specifically, the object will look ``smaller" if some rays erroneously carve into the object due to noise. Consequently, we optimize the consistency between the proposed mesh and the high-confidence cells in the vacancy volume, which we call ``high-confidence" surface points. We propose a global optimization to further refine the mesh based on these points. Recall that most rays intersect at the surface of the objects. Define point set $\hat{Y}$ as the point set of all high-confidence surface points of $V(x, y, z)$:
\begin{align}
\hat{Y} = \{(x, y, z) : V(x, y, z) > \epsilon_V \}
\end{align}
where $\epsilon_V$ is a threshold based on the carving statistics of volume $V$.
For a mesh reconstructed from running Marching Cubes, represented as a graph $G=(P, E)$, where $P$ is the set of vertices and $E$ is the set of edges that form the faces. A deformation function $f$ maps the original vertex set $P$ to a deformed set $P' = f(P)$. We first optimize a one-side Chamfer distance from the high-confidence surface points (less than $\epsilon$ distance away) to the mesh vertices. In addition, we regularize the mesh by a graph Laplacian loss. The final objective can be written as:
\begin{align*}
    L_{rf} &= \lambda_1 \frac{1}{|P'|} \sum_{{p'_i \in P' \atop ||p'_i-\hat{y}||_2 < \epsilon_d}} \min_{\hat{y} \in \hat{Y}} ||p'_i - \hat{y}||^2_2
               + \lambda_2 \frac{1}{|P'|} \sum_{p'_i \in P'} \sum_{p'_j \in \mathcal{N}(P')}^{} \frac{1}{\mathcal{N}(P'_i)} ||p'_j - p'_i||_2
\end{align*}
where $\mathcal{N}(P'_i)$ represents all neighbors of a node $P'_i$, and $\lambda_1$ and $\lambda_2$ are the weights between the two losses. We find a function $f$ that minimizes the loss. The $f$ function can be treated as the point-wise translation of the vertices. All values used in this optimization come from our predictions without using the ground truth. We use Adam Optimizer to optimize the warping function $f$.
        
\section{Experiments}
In this section, we present the data collection details, evaluation of the carving algorithm, and reconstruction of real objects.To better evaluate the performance of event-based 3D reconstruction algorithms, we collect Multi Object Event Camera Dataset in 3D (\dataset), a 3D event dataset of real objects. Please refer to the Supplementary Material for details about the dataset.
For ground truth models, an industrial-level Artec Spider scanner is used to provide the ground truth 3D models with high accuracy. The detailed steps of data collection can be found in the Supplementary Material.
\label{sec:data}
\begin{table}
    \begin{center}
    \caption{\textbf{Event Carving Evaluation} This table contains the results using ground truth Apparent Contour Events (ACEs). Chamfer distance (lower is better) is reported in $10^{-3} m$ (millimeters). Surface normal (higher is better) is reported as cosine similarity between the ground truth and predicted surface normal. We sample 10,000 points uniformly both on the reconstructed mesh and the object mesh.}
    \label{tab:synth}
        \begin{tabular}{c|c|c|c?c|c|c}
        \toprule
        \multicolumn{1}{c}{}& \multicolumn{3}{c?}{\textbf{Chamfer Distance$\downarrow$}} & \multicolumn{3}{c}{\textbf{Normal Consistency$\uparrow$}} \\
         Category & \ours{} & Mask-24 & Mask-12 & \ours{} & Mask-24 & Mask-12\\
          \hline
Mustard & \textbf{3.0164} & 4.6210 & 5.5161  & \textbf{0.9619} & 0.9034 & 0.9035\\
Coffee & \textbf{2.1439} & 2.2926 & 3.3019  & 0.9826 & \textbf{0.9893} & 0.9877\\
Soda (b) & 1.5231 & \textbf{1.4601} & 2.0635 & \textbf{0.9834} & 0.9717 & 0.9735 \\
Jello (s) & \textbf{0.9657} & 2.1973 & 4.3524  & \textbf{0.9801} & 0.9766 & 0.9234\\
Jello (b) & 5.9083 & \textbf{4.4409} & 7.7409  & 0.8843 & \textbf{0.9541} & 0.8952\\
Tuna & \textbf{3.2633} & 3.7045 & 4.3070 &  0.9598 & \textbf{0.9665} & 0.9644\\
Soup & \textbf{1.6513} & 2.0130 & 2.8556  & 0.9653 & \textbf{0.9705} & 0.9681\\
Sugar & \textbf{0.8651} & 2.4071 & 4.9491 & \textbf{0.9935} & 0.9862 & 0.9405\\
Vitamin & 2.6190 & \textbf{1.4478} & 2.4836 & 0.9683 & \textbf{0.9947} & 0.9896\\
Spam & \textbf{2.0398} & 3.1969 & 4.8615 & 0.9739 & \textbf{0.9760} & 0.9479\\
\hline
Mean & \textbf{2.4267} & 3.2652 & 4.3856 & \textbf{0.9487} & 0.9377 & 0.9159
    \end{tabular}
    \end{center}
\end{table}

\begin{figure}[t]
    \centering
        \includegraphics[width=.85\textwidth]{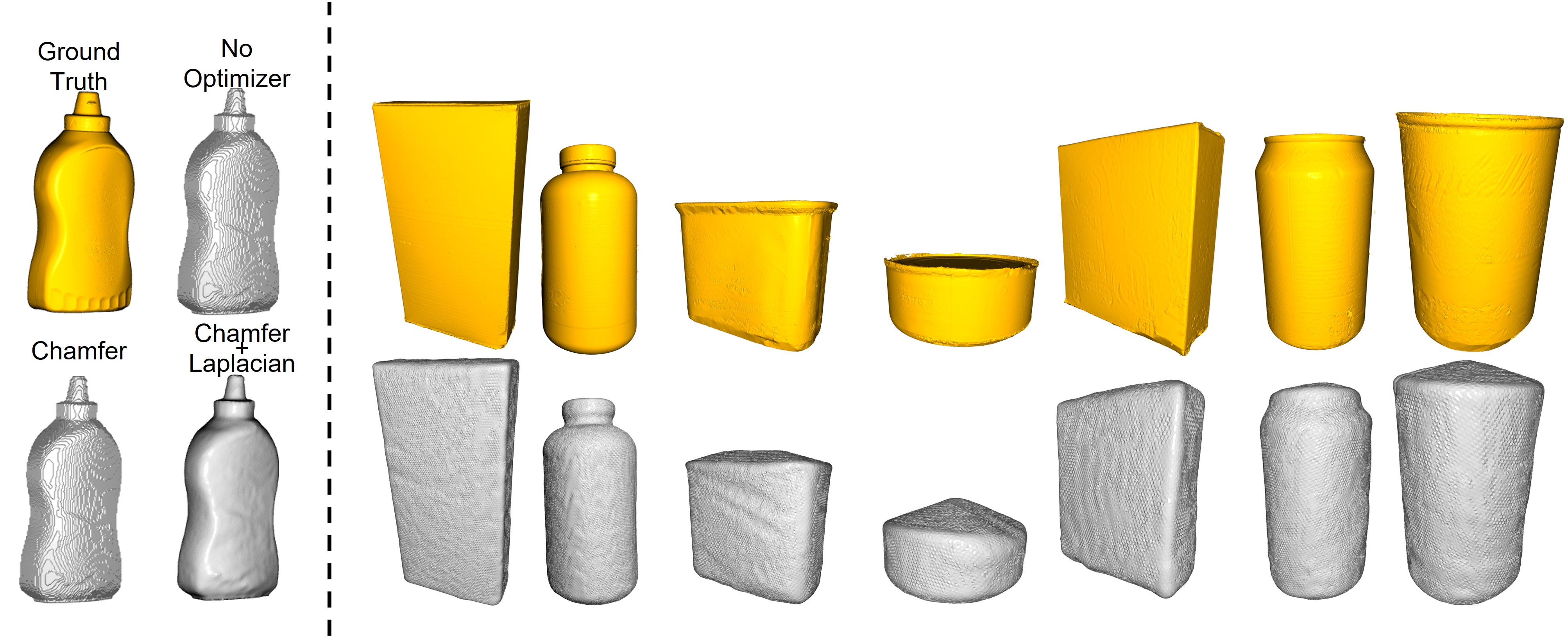}
 \caption{(Left) Comparison between options available within the optimizer. (Right) Test performance of event-based carving using predicted ACEs from our network. }
\label{fig:real_carving}
\end{figure}



\subsection{Evaluation Metrics} We report both Chamfer distance and Cosine similarity of the mesh compared to the ground truth model. Chamfer distance is measuring the average distance between two point clouds, which reflects the positional accuracy of the reconstruction. It is defined as:
\begin{align}
\small
\label{eqn:chamfer}
\begin{split}
CD(X, \hat{X}) = \frac{1}{|X|} \sum_{x \in X} \min_{\hat{x} \in \hat{X}} ||x - \hat{x}||_2 + \frac{1}{|\hat{X}|} \sum_{\hat{x} \in \hat{X}} \min_{\hat{x} \in \hat{X}} ||\hat{x} -x||_2
\end{split}
\end{align}
$X$ and $\hat{X}$ represent the points sampled from the reconstruction and the ground truth model. Surface normal is also a commonly used metric for comparing the geometry of two meshes. We report the average cosine similarity between the corresponding surface samples of two meshes, which is defined as:
\begin{align}
&Cos.Sim(X_{gt}, X_{pred}) = \frac{1}{|X_{gt}|}\sum_{i \in |X_{gt}|}{|{\Vec{n_i}\cdot{\Vec{m}_{\theta(x, X_{pred})}}}|} \\
&\theta(x, X_{gt}\coloneqq\{(\Vec{y_j}, \Vec{m_j})\})) = \argmin_{j \in |X_{gt}|} ||x - y_j||_2^2 \end{align}
We use the closest point to approximate the correspondence between two sets of oriented samples, similar to the argmin used in Equation~\ref{eqn:chamfer}. We use a k-nearest neighbor search to estimate the normals of sampled points on the mesh, where $k$ is $300$.

\newcolumntype{L}{>{\centering\arraybackslash}p{1.20cm}}
\newcolumntype{K}{>{\centering\arraybackslash}p{1.10cm}}
\newcolumntype{J}{>{\centering\arraybackslash}p{1.00cm}}

\begin{table}[t]
    \centering
    \caption{\textbf{Real Object Reconstruction} This table contains the results using trained network to predict Active Contour Events from real data. Chamfer distance (lower is better) is reported in $10^{-3} m$ (millimeters). Surface normal (higher is better) is reported as cosine similarity between the ground truth and predicted surface normal. ``Mask" means using masks from the event mask network. ``Image" means using masks predicted from reconstructed images. The number in each column name represents the number of views used for reconstruction.}
    \label{tab:real}
        \begin{tabular}{c|c|c|c|c|c|c}
        \toprule
        \multicolumn{1}{c}{}& \multicolumn{6}{c}{\textbf{Chamfer Distance$\downarrow$ / Surface Normal Consistency$\uparrow$}}\\
        
          & \ours{} & \shortstack{Mask 24} & \shortstack{Mask 12} & \shortstack{Image 24}& \shortstack{Image 12} & \shortstack{E3D~\cite{baudron2020e3d}}\\
          \hline
          Mus   & \textbf{1.537}/\textbf{0.983} & 3.034/0.968            & 7.061/0.926       & 4.192/0.868                       & 6.947/0.926 & 7.986/0.713\\
          Cof   & \textbf{2.286}/0.957 & 2.653/\textbf{0.971}            & 7.771/0.915        & 5.733/0.840                      & 5.930/0.821 & 8.354/0.756\\
          Sod   & 2.239/\textbf{0.965}         & \textbf{1.953}/0.957    & 4.611/0.929          & 2.380/0.914                    & 3.865/0.884 & 6.762/0.703\\
          Jel(s) & \textbf{2.889}/\textbf{0.928} & 3.860/0.925            & 14.248/0.783       & 3.967/0.862                     & 7.188/0.757 & 8.255/0.744\\
          Jel(b) & \textbf{3.899}/\textbf{0.930}          & 4.818/0.926   & 13.405/0.750       & 3.929/0.863            & 6.657/0.736 &14.910/0.767\\
          Tun   & \textbf{3.624}/\textbf{0.938} & 3.518/0.937            & 5.552/0.753          & 4.254/0.863                    & 8.545/0.734 &10.850/0.704\\
          Sou   & \textbf{2.111}/\textbf{0.959} & 2.392/0.954           & 5.200/0.887          & 2.294/0.922                    & 5.276/0.854 & 6.133/0.783\\
          Sug   & \textbf{1.953}/\textbf{0.970}          & 7.904/0.854            & 9.929/0.833         & 4.000/0.939   & 9.724/0.775 & 5.924/0.691\\
          Vit   & \textbf{2.191}/0.957 & 2.338/\textbf{0.966}            & 5.772/0.949       & 2.226/0.958                       & 5.710/0.915 & 8.462/0.715\\
          Spa   & \textbf{2.784}/\textbf{0.953} & 3.667/0.945            & 9.635 /0.738         & 3.295/0.911                   & 6.798/0.849  &10.730/0.747\\
          \hline
          Mean      & \textbf{2.551}/\textbf{0.954} & 3.614/0.940 & 8.312/0.846 &  3.602/0.900 & 6.664/0.825 & 8.837/0.732
        \end{tabular}
\end{table}

\subsection{Evaluating Carving Algorithm}
To test the effectiveness of our continuous carving algorithm, we utilize the meshes collected as part of this dataset within a simulation environment for fair comparisons. Note that the evaluation is done with real objects and we assume the ACEs are known at every point during the camera motion. This is different than the completely synthetic environment employed in \cite{baudron2020e3d} because the events generated with an event simulator are not guaranteed to have the same data distribution. We observe a significant amount of noise in the real event data. The quantitative results are provided in Table~\ref{tab:synth}. The mask-based carving is done with ground truth masks as well for fair comparison.

In addition to the real data simulation above, we show ShapeNet examples, Figure~\ref{fig:shapnet_results}, of our algorithm on objects with more complicated geometry to provide context for our reconstruction quality. To use these models, we use Open3D~\cite{Zhou2018} to capture high frame rate images and ground truth masks. These images were then processed through ESIM~\cite{Gehrig_2020_CVPR} to generate a set of simulated events. To generate a close approximation of the real world dataset, a similar trajectory to the real world dataset was chosen. 
\subsection{Reconstructing Real Objects}
Many network-based methods only work on simulated datasets because they require a large amount of labeled object-level 3D models. In addition, such networks cannot easily be adapted to work on real data. In comparison, the \ours{} network can be trained on a small set of data because the labels could be obtained geometrically for events. We report the per-class performance evaluation in Table~\ref{tab:real}. For each object, we evaluate on an unseen sequence withheld from the training set. \ours{} uses apparent contour events from the network output to perform carving. For baseline comparisons, we train two separate U-Net~\cite{ronneberger2015u} style networks to output object masks from previous events and from reconstructed images. While they differ in input, they emulate the common situations where a fixed number of frames are used for reconstruction. 

We follow the multi-view settings in 3D-R2N2~\cite{choy20163d} where views are taken around the object. We choose 12 views as the baseline because we can reconstruct reasonable objects while keeping the computational cost close to \ours{}. To further show the computational efficiency of \ours{}, we also compare with 24-view carving, whose computational cost is much higher. We compare with E3D~\cite{baudron2020e3d}, the only event-based method that attempts to achieve multi-view 3D reconstruction. For fair comparison, we directly feed in the ground truth poses to E3D. E3D uses multi-view silhouette optimization over the objects, similar to PMO~\cite{lin2019photometric}. E3D directly uses the photometric optimization module in PMO~\cite{lin2019photometric} on silhouettes and removes the mesh prior from AtlasNet~\cite{groueix2018papier}. In our evaluation, we feed ground truth poses to E3D for fair comparison. In our experiments, we find silhouette-based optimization methods sensitive to the position and size of the mesh.
%
To study the various components of \ours{}, we report the performance of the ACE classification network and overall object reconstruction. For ACE classification, we provide the AUC curve of the classifier in Fig.~\ref{fig:sphere_and_auc} (b). The overall classification accuracy is 0.9563 (threshold=0.5). In Table~\ref{tab:compute}, we show the mean performance and the mean number of operations. We define number of operations as the number of rays that we shoot out of the camera. \ours{} uses significantly fewer operations than both 12 and 24 views. 
We notice that for both mask prediction networks, the quality of reconstruction degrades quickly when the number of views decreases. In practice, the sensor frame rate is not the only limiting factor - the computational power required to carve based on masks is also significantly higher. The average number of carving operations, mean Chamfer distance, and mean normal consistency are summarized in Table~\ref{tab:compute}. This means the reconstruction quality of a frame-based algorithm largely depends on the motion speed, assuming the camera sensor has a fixed frame rate. We overcome this limitation of motion speed by directly operating on a continuous stream of events. We directly compare the qualitative results of the discussed methods in Figure~\ref{fig:evac_carving}. 
\begin{table}[t]
    \centering
    \caption{Mean number of carving operations, mean Chamfer distance, and mean cosine similarity. With ACEs, our continuous carving method outperforms the other frame-based methods while using significantly fewer operations.}
    \label{tab:compute}
    \begin{tabular}{c|c|c|c}
    \toprule
    Method & Num of Ops$\downarrow$ & Chamfer$\downarrow$ & Normal $\uparrow$ \\
    \hline
        GT-Mask-24 & 6,661,111 & 3.614 & 0.940 \\
        GT-Mask-12 & 3,331,536 & 8.312 & 0.846 \\
        Image-24 & 6,674,148 & 3.602 & 0.900 \\
        Image-12 & 3,345,580 & 6.664 & 0.825 \\
        E3D~\cite{baudron2020e3d} & -- & 8.837 & 0.732\\
        \ours{}  & \textbf{1,921,976} & \textbf{2.551} & \textbf{0.954} \\
    \hline
    \end{tabular}
\end{table}

\subsection{Real Objects with Handheld Camera Trajectory}
\begin{figure}
    \centering
    \includegraphics[width=\textwidth]{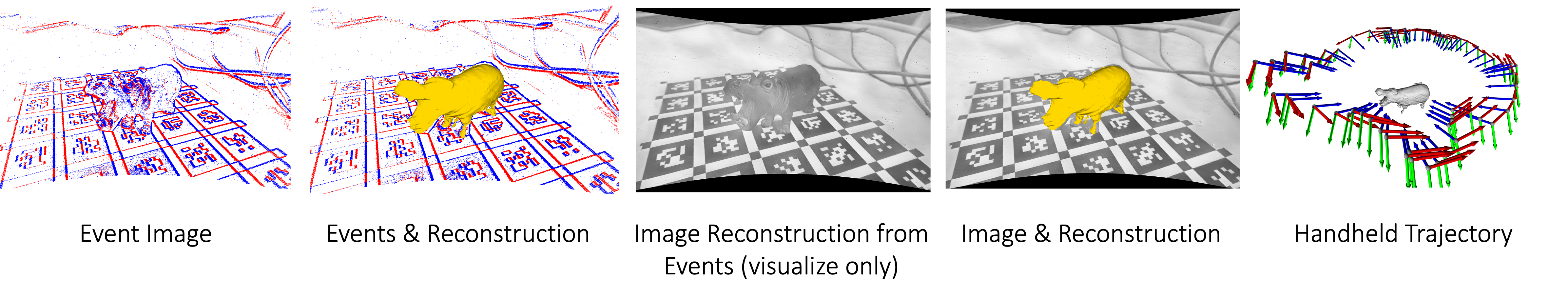}
    \caption{Results from a handheld trajectory. Left to right: raw events input, raw events overlaid with our reconstruction, image reconstruction using E2Vid~\cite{rebecq2019high}, image reconstruction overlaid with our reconstruction, and the subsampled 3D camera trajectory with the computed mesh.}
    \label{fig:handheld}
\end{figure}

In the previous section, we present the experimental results for circular trajectories. However, camera trajectories can have more degrees of freedoms in real life. In this section, we put \ours{} under test of more general handheld motions. The additional complexity of tasks comes not only from significant background events, but also the noisy camera pose estimation from handheld camera motion. 
We show a reconstructed hippo in Figure~\ref{fig:handheld}.
Our reconstruction on this handheld sequence shows success in the main body of the hippo with an average reconstruction error of 1.5mm. The legs do not appear fully formed likely due to the small errors in the calibration and pose, both of which rely upon the reconstructed image to detect the AprilTags.
\section{Conclusions}
In this work, we present a novel method for continuous 3D reconstruction using event cameras. At the core of the method is the representation of occluding contours by Apparent Contour Events (ACE), a novel event quantity that can be used to continuously carve out high-fidelity meshes. \ours{} is able to update the occupancy grid of the object on an event-to-event basis, which achieves better performance than mask-based visual hull approaches while using significantly fewer carving operations. We evaluate the performance of the method on both real and synthetic data. In addition, we contribute \dataset{}, the first high-quality event-based 3D object dataset. With these contributions, we believe \ours{} can provide important insights into how we can understand the 3D world through events.

\textbf{Acknowledgement} We thank the support from the following grants: NSF TRIPODS 1934960, NSF CPS 2038873, ARL DCIST CRA W911NF-17-2-0181, ARO MURI W911NF-20-1-0080, ONR N00014-17-1-2093, DARPA-SRC C-BRIC, and IARPA ME4AI. We also thank William Sturgeon from the Fisher Fine Arts Materials Library for providing the Artec Spider scanner and assistance.

%
%
\bibliographystyle{splncs04}
\bibliography{egbib}
\end{document}